\pdfoutput=1

\documentclass[11pt]{article}

\usepackage{acl}

\usepackage{times}
\usepackage{latexsym}
\usepackage{amsmath}
\usepackage{enumitem}
\usepackage{alphabeta}
\usepackage{comment}
\usepackage[font=small,skip=10pt]{caption}

\usepackage{subfig}
\usepackage{appendix}
\usepackage{blindtext}
\usepackage{graphicx}

\usepackage[T1]{fontenc}

\usepackage[utf8]{inputenc}

\usepackage{microtype}

\title{k-Rater Reliability:\\The Correct Unit of Reliability for Aggregated Human Annotations

\normalsize{ACL 2022 Main Conference}}

\author{Ka Wong \\
  Google Research \\
  \texttt{danicky@gmail.com} \\\And
  Praveen Paritosh \\
  Google  Research\\
  \texttt{pkp@google.com} \\}

\begin{document}
\maketitle
\begin{abstract}
Since the inception of crowdsourcing, aggregation has been a common strategy for dealing with unreliable data. Aggregate ratings are more reliable than individual ones. However, many natural language processing (NLP) applications that rely on aggregate ratings only report the reliability of individual ratings, which is the incorrect unit of analysis. In these instances, the data reliability is under-reported, and a proposed \textit{k-rater reliability} (kRR) should be used as the correct data reliability for aggregated datasets. It is a multi-rater generalization of \textit{inter-rater reliability} (IRR). We conducted two replications of the WordSim-353 benchmark, and present empirical, analytical, and bootstrap-based methods for computing kRR on WordSim-353. These methods produce very similar results. We hope this discussion will nudge researchers to report kRR in addition to IRR.
\end{abstract}

\section{Introduction}
Crowdsourcing has become a mainstay for data collection in NLP \cite{geva2019, sabou2014}. It can produce data in a scalable and cost effective manner. However, these benefits come at a cost: quality. The reliability of crowd workers is always of central concern. One common strategy to increase the data reliability is to collect multiple, independent judgements and to use the aggregated judgements instead. Indeed, early papers such as \citet{snow2008} show that average ratings correlate more strongly with expert judgements. This makes sense, as average ratings are known to have a higher reliability than individual ones \cite{ebel1951}.

A number of strategies have been proposed to address data quality issues, e.g. rater modeling, label correction, label pruning \cite{kumar2011}, but aggregation remains very popular \cite{prabhakaran2021}. \citet{sheshadri2013square} present nine crowdsourced datasets across a wide range of NLP tasks to compare different aggregation methods. See \citet{difallah2021aggregation} for a recent review of aggregation techniques. In short, aggregation has become the default method for acquiring reliable data from the crowd.

Interestingly, after we adopted aggregation as a community, we forgot to update our reliability measures correspondingly. The field continues to report data reliability in terms of IRR, even when aggregate ratings are used. Focusing on IRR, we are unable to capture the increase in reliability due to aggregation. The actual data reliability is hence unknown. This has important consequences. Reliability is often used as a safeguard for \textit{reproducibility}. Therefore conclusions about the reproducibility of a dataset drawn based the reliability of individual ratings may be different than that based on the reliability of aggregate ratings.

By reporting the correct reliability that is actually higher, this may even have a side effect of lessening the stigma on low-IRR datasets. As a result, this may create a path forward towards reliable data on subjective tasks, where a high IRR is difficult to obtain, such as emotions \cite{wong2021} and toxicity \cite{wulczyn2017ex}. With a reproducibility crisis looming in the background 
\cite{baker2016reproducibility, hutson2018artificial}, more frequent and accurate reporting of reliability is our primary safeguard \cite{Paritosh2012}.

We denote the reliability of aggregate ratings as \textit{$k$-rater reliability} (kRR), in order to differentiate it from inter-rater reliability. In this paper we present a few methods for computing kRR. First, we demonstrate a general, empirical approach that is based on replications. To that end, we conducted two replications of WordSim-353 \cite{finkelstein}, a widely used word similarity dataset. We then discuss two other alternatives that do not require replications. One is a re-sampling-based bootstrap approach \cite{efron1994}. It is suitable for experiments with a high rating redundancy.  The other is an existing analytical approach based on intraclass correlation (ICC). It is suitable for continuous data where the aggregation is the mean. We conclude with recommendations for reporting reliability of crowdsourced annotations, and novel research questions to expand the usefulness of kRR.

\section{Related Work}
Various authors have stressed the importance of measuring reliability for the correct unit of analysis. \citet{ebel1951} asks ``Is it better to estimate the reliability of individual ratings or the reliability of average ratings? If decisions are based upon average ratings, it of course follows that the reliability with which one should be concerned is the reliability of those averages.'' \citet{shrout1979} and \citet{Hallgren2012} reiterate similar points.

These studies primarily focus on the reliability of the \textit{mean}, which is just one of many different aggregation methods. There is a reason. Not only is the mean a popular choice, it is also the only known choice where the reliability of the aggregate ratings can be computed \textit{analytically} from the reliability of individual ratings. This is done in the ICC framework. ICC is typically used to measure the reliability of single ratings, but it actually has a variant that can be used for mean ratings as well. \citet{shrout1979} list several types of ICC coefficient, one of which is for mean ratings. They call it ICC($k$), where $k$ is the number of ratings per item. In this generalized notation, ICC(1) is just the reliability of individual ratings, or the IRR. Note that \citet{mcgraw1996} use a slightly different notation, ICC($1,k$), to explicitly denote that it is for a one-way random effects model, where the raters are treated as interchangeable. That is a common assumption in most crowdsourcing experiments done on commercial platforms such as Amazon Mechanical Turk.

ICC($k$) is an established way of measuring the reliability of mean ratings, hence it is readily usable by researchers. However, it has some drawbacks. Being part of the ICC family, ICC($k$) is only applicable to continuous data. In addition, ICC($k$) measures the reliability of \textit{mean} ratings, therefore it cannot accommodate other aggregation functions. In other words, for other popular data types, such as majority votes of binary data, there is no known coefficients for measuring the reliability of aggregate ratings. Other than ICC($k$), the authors are not aware of any multi-rater generalization for other coefficients such as Cohen's (\citeyear{cohen1960}) \textit{kappa} or Krippendorff's \textit{alpha} \cite{kripp2011}. We therefore take ICC($k$) as an inspiration and abstract away from it to define a class of reliability that describes the reliability of aggregate ratings for any data types. We denote it kRR.

\section{Contributions}
\begin{itemize}
  \item We emphasise the reliability of aggregate ratings is higher than that of individual ratings. 
  \item We give a general definition of kRR, extending from the definition of IRR, and discuss three methods for computing it.
  \item We conduct two replications of the WordSim-353 benchmark to validate these methods.
\end{itemize}

\section{\textit{k}-Rater Reliability}
\label{sec:kRR}

We define kRR as the chance-adjusted agreement between replications of aggregate ratings. This definition is very similar to IRR. In fact, they only differ in terms of interpretation. kRR is identical to IRR other than that each individual rating in the IRR calculation is replaced by a $k$-rater aggregate rating. After all, the mathematics in IRR are agnostics to how those labels are produced.

Just like IRR, a minimum of two replications is required to calculate kRR. Given two vectors of aggregate ratings, one can calculate the reliability between them using any IRR coefficients that fit the purpose. kRR is designed to be analogous to IRR so that we can build upon the rich IRR literature and the various coefficient choices for different experimental conditions and assumptions. For example, in a binary task, if all the items are rated by two fixed but distinct groups of raters (raters from different locales), Cohen's (\citeyear{cohen1960}) \textit{kappa} is a suitable choice. Whereas if the raters groups are homogeneous, and the rating scale is ordinal (e.g. Likert), then Krippendorff's \textit{alpha} \cite{kripp2011} can be used. Just like IRR, kRR is a general concept and is agnostic to the choice of coefficient.

This definition of kRR can be directly operationalized by creating replications. We call this approach to calculating kRR the \textit{empirical approach}. We demonstrate it in the next section on the WordSim-353 benchmark. The empirical approach is the most direct and most general, with the drawback that a minimum of two replications are required. We later present two narrower alternatives in Section \ref{sec:alternatives} that do not require replications. The empirical results will be used as a golden reference to validate them.

\subsection{Replicating the WordSim Dataset}
\label{sec:rep_wordsim}

WordSim-353 \cite{finkelstein} is a widely used benchmark for measuring a system’s ability to compute similarity between two words, and has been cited over 1500 times. The dataset contains 353 word pairs. Each word pair is rated by the same 13 workers for their similarity on a scale from 1 to 10, to indicate how similar their meanings are. The 13 ratings on each word pair are then aggregated into a mean score. It is important to note that only the mean of the ratings are utilized by all the research using this dataset as a \href{https://aclweb.org/aclwiki/WordSimilarity-353_Test_Collection_(State_of_the_art)}{benchmark}. So the unit of analysis is the aggregate of the 13 ratings, not individual ratings. 

Nearly twenty years have elapsed since the creation of the WordSim dataset. It is impossible to recreate the original experimental conditions due to rater population changes. Therefore, we created two replications in order to approximate the kRR of the original dataset. Two is the minimum replication factor required for the empirical approach, though a higher replication would result in a more accurate measure of kRR. 

We used the original annotation guidelines on Amazon Mechanical Turk. Raters were paid on average USD 9.5 per hour. In each replication, we collected 13 judgements on each of the same 353 word pairs. There was a detail that we did not follow. In the original experiment, the authors employed 13 unique raters, and each one rated all 353 word pairs. In our replications, we followed more modern conventions and limited the contributions of each individual rater for better generalizability. This detail aside, these are our best attempts to replicate the original experiment. The data is publicly available at \url{https://github.com/google-research-datasets/wordsim-replications}.

\subsection{Empirical kRR Results}
\label{sec:kRR_empirical}
\begin{figure}
  \centering
  \includegraphics[width=0.48\textwidth]{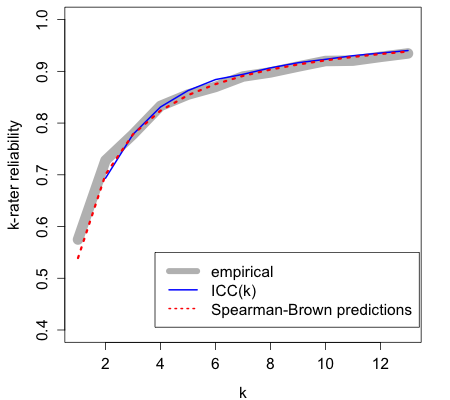}
  \caption{$k$-rater reliability for replications of WordSim benchmark, calculated using 3 different methods: 1) Empirical, based on replications, 2) ICC($k$), analytical, and 3) SB predictions. Note ICC(1) is not available as we only have a single column of ratings available at $k=1$. All SB predictions are based on only 2 ratings per item.}
  \label{fig:3in1}
\end{figure}

We take $k$ columns of ratings at random from each of the two replications, compute the $k$-rater mean scores for each replication, and measure the reliability between them using Krippendorf's \textit{alpha}, the most widely used and general reliability index. We do this for $k=1,2,\ldots,13$. The resulting kRR values are shown in Fig.\ref{fig:3in1}. At $k=1$, the IRR is 0.574, slightly lower than the 0.6 originally reported in \citet{finkelstein}. At $k=13$, the $k$-rater reliability is 0.940, quite a bit higher than the IRR. In addition, Fig.\ref{fig:3in1} shows the marginal returns on increasing the number of ratings on the replicated datasets.

\section{Other Approaches to Computing kRR}
\label{sec:alternatives}
The empirical approach is general, as it can accommodate any choice of rating scale, aggregation function, and reliability coefficient. However, it has a major drawback. As we see in Section \ref{sec:rep_wordsim}, it can be difficult to do a perfect replication post-fact. This backward incompatibility will present a challenge to computing kRR for existing datasets. Below we present two alternatives that can work on existing datasets under some conditions without requiring any additional data collection. One is a re-sampling based bootstrap approach \cite{efron1994}, the other is ICC($k$).

\subsection{Bootstrap}
\label{sec:bootstrap}
Bootstrap \cite{efron1994} is a re-sampling technique commonly used for quantifying uncertainty in statistical parameter estimation. One can bootstrap an NLP annotations dataset by re-sampling ratings within each annotation item with replacement at the same sample size. If one treats each bootstrap sample as a replication, then one can apply the technique discussed in Section \ref{sec:kRR} to obtain a \textit{bootstrapped} kRR. Bootstrap is an approximate technique and works better with larger sample sizes, typically 20 observations and above for a single distribution. The 13-rating redundancy in the WordSim replications is arguably small for a typical bootstrap exercise, but it makes up for it with a large number of items.

Before we apply bootstrap to the original WordSim dataset, we first verify its soundness by comparing it against the empirical results in Section \ref{sec:kRR_empirical}. When applied to one of the two recent replications, the bootstrapped kRR is 0.943. This is comparable to the 0.940 reported in Section \ref{sec:kRR_empirical}. We then apply bootstrap to the original WordSim dataset and find a bootstrapped kRR of 0.953 (Table \ref{tab:ICC}). The exact method introduced below produces a very similar value at 0.950.

\subsection{Intraclass Correlation}
Intraclass correlation is a popular reliability coefficient for continuous data in behavioral and medical sciences. ICC gives researchers granular control over assumptions about the raters. For example, each annotation item can be rated by the same set of raters, or different sets of raters (interchangeability). In the former, the raters can be treated as either fixed or randomly drawn from a population. \citet{shrout1979} and \citet{mcgraw1996} give very extensive treatment on different ICC types for different rater assumptions.

In this paper, we focus on the most basic definition, one that treats raters as interchangeable. The ICC for $k$-rater averages is denoted as ICC($k$) using  \citeauthor{mcgraw1996}'s notation. The reliability of individual ratings is thus given by ICC(1). ICC($k$) can be computed by summing squares of differences on the data matrix. Please see Appendix~\ref{appendix} for derivation and an illustration. Otherwise, software implementations of ICC are also widely available, e.g. in R and Python.

We first verify ICC($k$)'s accuracy by comparing it against the empirical results in Section \ref{sec:kRR_empirical}. To do that, we calculate ICC($k$) for one of the two recent WordSim replications for $k=1,2,\ldots,13$ and overlay the results (solid blue) over the empirical curve in Fig.\ref{fig:3in1}. We can see ICC($k$) matches the empirical results quite well.


After verifying the technique, we compute ICC($k$) on the original WordSim dataset. We report in Table \ref{tab:ICC} both ICC(1) and ICC(13) to show the increase in reliability. They are respectively 0.590 and 0.950.\footnote{The former is computed using two-way random without interaction ICC(1), the latter two-way random without interaction ICC(13). The equivalent one-way models yield identical point estimates.}

\begin{table}
 \centering
   \begin{tabular}{|c c c|}
    \hline
    \textbf{Unit of analysis} & \textbf{Method} & \textbf{reliability}\\
    \hline
    single-rating & ICC(1) & 0.590\\
    \hline
    13-rating mean & ICC(13) & 0.950\\
    \hline
    13-rating mean & bootstrap & 0.953\\
    \hline
   \end{tabular}
 \caption{\label{tab:ICC} Reliability of the original WordSim benchmark. First two rows are analytical estimates ICC(1) and ICC(13). Both computed using all 13 available ratings. Third row is a re-sampling-based bootstrapped estimate based on 100 bootstrap samples.}
\end{table}

\subsection{Spearman-Brown Formula}
Given an experiment with a $k$-rating redundancy, ICC($k$) quantifies the reliability of the $k$-rater average. If this reliability is too low, the researcher may want to increase the value of $k$. In this case, it would be helpful to know how additional ratings would impact reliability. This is analogous to calculating the required sample size for a given margin of error in a poll. For this purpose, the Spearman-Brown prophecy formula \cite{spearman1910, brown1910} can be a useful tool. It predicts ICC($k$) for any value of $k$ based on ICC(1) in the current experiment:
\begin{equation} \label{eq:SB}
\textrm{ICC($k$)}=\frac{k \cdot \textrm{ICC(1)}}{1+(k-1) \cdot \textrm{ICC(1)}}.
\end{equation}
\citet{warrens2017} and \citet{devet2017} recently proved that SB and ICC($k$) are indeed equivalent in expectation,\footnote{The only exception is two-way mixed model with interaction \cite{warrens2017}.} even though they look nothing alike and were derived in very different contexts. These findings confirm past observations that SB predicts empirical results accurately \cite{remmers1927}. A limitation of SB is clearly that it only works with ICC. However, \citet{fleiss1973} show ICC is actually equivalent to weighted-kappa with quadratic weights, so it likely has wider applicability.

To verify the formula, we apply SB to one of the two recent WordSim replications and overlay the results (dotted red) over the empirical curve obtained earlier. When computing SB, we only provide it with 2 ratings, in order to assess its predictive accuracy. That is, we first compute ICC(1) with 2 randomly drawn ratings from each word pair, then we plug this ICC(1) value into Eq.\ref{eq:SB} for $k=1,2,\ldots,13$. The SB curve is overlaid over the empirical curve in Fig.\ref{fig:3in1}. We see that SB tracks the empirical results very well even at high $k$. This is remarkable as the empirical approach requires 26 ratings for $k=13$, whereas SB merely requires 2 for any value of $k$.


\section{Conclusions and Discussion}
 
We pointed out where aggregated ratings are used, as is the case in many crowdsourced datasets, reliability of aggregate ratings is the correct accounting of data reliability. We introduced $k$-rater reliability (kRR) as a multi-rater extension of IRR. We emphasise the reliability of aggregate ratings is higher than that of individual ratings. We present analytical and bootstrap-based methods for computing the kRR on the original WordSim dataset. Both methods produce similar estimates for 13-rater reliability ranging from 0.940 to 0.953. We conduct two replications of the entire WordSim-353 benchmark to validate these methods. We make our replication data publicly available on GitHub.

While aggregation makes it possible to have reliable benchmarks on subjective topics, some readers may feel uneasy about increasing reliability via gathering additional ratings, as opposed to other traditional means such as improving rater guidelines. We suggest to mediate this concern by reporting both IRR and kRR. In fact, kRR is not meant to replace IRR, but rather complement it. IRR speaks to the reliability of the labeling process, whereas kRR quantifies the reliability of the aggregated data we consume. We urge researchers to report both where possible. In fact, \citet{Hallgren2012} states, "In cases where single measures ICCs are low but average-measures ICCs are high, the researcher may report both ICCs to demonstrate this discrepancy."

This research also raises interesting questions for future research:
\begin{enumerate}
    \item How do we derive multi-rater generalizations for coefficients other than ICC? A lot of NLP annotations are binary and multi-class. Such a generalization for majority voting would be particularly useful to the field.
    
    \item Should we apply the \citet{Landis:1977} style of reliability cutoffs to kRR, or should kRR go by a different set of standards?
\end{enumerate}

We urge researchers to report both IRR and kRR of aggregated human annotations, and for further inquiry around the above fundamental questions about reliability.

\section*{Acknowledgement}
We thank Lora Aroyo and Chris Welty for sharing their WordSim replication datasets. We thank Michael Quinn and Jeremy Miles for their insightful discussions and comments. We also thank all the crowd workers for providing us with valuable annotations data. 







\bibliographystyle{acl_natbib}
\bibliography{kRR}

\begin{thebibliography}{28}
\expandafter\ifx\csname natexlab\endcsname\relax\def\natexlab#1{#1}\fi

\bibitem[{Baker(2016)}]{baker2016reproducibility}
Monya Baker. 2016.
\newblock Reproducibility crisis.
\newblock \emph{Nature}, 533(26):353--66.

\bibitem[{Brown(1910)}]{brown1910}
William Brown. 1910.
\newblock Some experimental results in the correlation of mental abilities 1.
\newblock \emph{British Journal of Psychology, 1904-1920}, 3(3):296--322.

\bibitem[{Cohen(1960)}]{cohen1960}
Jacob Cohen. 1960.
\newblock A coefficient of agreement for nominal scales.
\newblock \emph{Educational and psychological measurement}, 20(1):37--46.

\bibitem[{{de Vet} et~al.(2017){de Vet}, Mokkink, Mosmuller, and
  Terwee}]{devet2017}
Henrica~C.W. {de Vet}, Lidwine~B. Mokkink, David~G. Mosmuller, and Caroline~B.
  Terwee. 2017.
\newblock \href
  {https://doi.org/https://doi.org/10.1016/j.jclinepi.2017.01.013}
  {Spearman-brown prophecy formula and cronbach's alpha: different faces of
  reliability and opportunities for new applications}.
\newblock \emph{Journal of Clinical Epidemiology}, 85:45--49.

\bibitem[{Difallah and Checco(2021)}]{difallah2021aggregation}
Djellel Difallah and Alessandro Checco. 2021.
\newblock Aggregation techniques in crowdsourcing: Multiple choice questions
  and beyond.
\newblock In \emph{Proceedings of the 30th ACM International Conference on
  Information \& Knowledge Management}, pages 4842--4844.

\bibitem[{Ebel(1951)}]{ebel1951}
Robert~L Ebel. 1951.
\newblock Estimation of the reliability of ratings.
\newblock \emph{Psychometrika}, 16(4):407--424.

\bibitem[{Efron and Tibshirani(1994)}]{efron1994}
Bradley Efron and Robert~J Tibshirani. 1994.
\newblock \emph{An introduction to the bootstrap}.
\newblock CRC press.

\bibitem[{Finkelstein et~al.(2001)Finkelstein, Gabrilovich, Matias, Rivlin,
  Solan, Wolfman, and Ruppin}]{finkelstein}
Lev Finkelstein, Evgeniy Gabrilovich, Yossi Matias, Ehud Rivlin, Zach Solan,
  Gadi Wolfman, and Eytan Ruppin. 2001.
\newblock Placing search in context: The concept revisited.
\newblock In \emph{Proceedings of the 10th international conference on World
  Wide Web}, pages 406--414.

\bibitem[{Fleiss and Cohen(1973)}]{fleiss1973}
Joseph~L Fleiss and Jacob Cohen. 1973.
\newblock The equivalence of weighted kappa and the intraclass correlation
  coefficient as measures of reliability.
\newblock \emph{Educational and psychological measurement}, 33(3):613--619.

\bibitem[{Geva et~al.(2019)Geva, Goldberg, and Berant}]{geva2019}
Mor Geva, Yoav Goldberg, and Jonathan Berant. 2019.
\newblock Are we modeling the task or the annotator? an investigation of
  annotator bias in natural language understanding datasets.
\newblock \emph{arXiv preprint arXiv:1908.07898}.

\bibitem[{Hallgren(2012)}]{Hallgren2012}
Kevin~A Hallgren. 2012.
\newblock \href {https://doi.org/10.20982/tqmp.08.1.p023} {Computing
  inter-rater reliability for observational data: An overview and tutorial.}
\newblock \emph{Tutorials in quantitative methods for psychology}, 8(1):23--34.

\bibitem[{Hutson(2018)}]{hutson2018artificial}
Matthew Hutson. 2018.
\newblock Artificial intelligence faces reproducibility crisis.

\bibitem[{Krippendorff(2011)}]{kripp2011}
Klaus Krippendorff. 2011.
\newblock Computing krippendorff's alpha-reliability.

\bibitem[{Kumar and Lease(2011)}]{kumar2011}
Abhimanu Kumar and Matthew Lease. 2011.
\newblock Modeling annotator accuracies for supervised learning.
\newblock In \emph{Proceedings of the Workshop on Crowdsourcing for Search and
  Data Mining (CSDM) at the Fourth ACM International Conference on Web Search
  and Data Mining (WSDM)}, pages 19--22.

\bibitem[{Landis and Koch(1977)}]{Landis:1977}
J.~Richard Landis and Gary~G. Koch. 1977.
\newblock \href {https://www.jstor.org/stable/2529310?seq=1} {The measurement
  of observer agreement for categorical data.}
\newblock \emph{Biometrics}, 33(1):159--74.

\bibitem[{Liljequist et~al.(2019)Liljequist, Elfving, and
  Skavberg~Roaldsen}]{liljequist2019}
David Liljequist, Britt Elfving, and Kirsti Skavberg~Roaldsen. 2019.
\newblock Intraclass correlation--a discussion and demonstration of basic
  features.
\newblock \emph{PloS one}, 14(7):e0219854.

\bibitem[{McGraw and Wong(1996)}]{mcgraw1996}
Kenneth~O McGraw and Seok~P Wong. 1996.
\newblock Forming inferences about some intraclass correlation coefficients.
\newblock \emph{Psychological methods}, 1(1):30.

\bibitem[{Paritosh(2012)}]{Paritosh2012}
Praveen Paritosh. 2012.
\newblock \href {https://research.google/pubs/pub40741/} {Human computation
  must be reproducible}.
\newblock In \emph{WWW 2012, Lyon.}

\bibitem[{Prabhakaran et~al.(2021)Prabhakaran, Mostafazadeh~Davani, and
  Diaz}]{prabhakaran2021}
Vinodkumar Prabhakaran, Aida Mostafazadeh~Davani, and Mark Diaz. 2021.
\newblock \href {https://aclanthology.org/2021.law-1.14} {On releasing
  annotator-level labels and information in datasets}.
\newblock In \emph{Proceedings of The Joint 15th Linguistic Annotation Workshop
  (LAW) and 3rd Designing Meaning Representations (DMR) Workshop}, pages
  133--138, Punta Cana, Dominican Republic. Association for Computational
  Linguistics.

\bibitem[{Remmers et~al.(1927)Remmers, Shock, and Kelly}]{remmers1927}
HH~Remmers, NW~Shock, and EL~Kelly. 1927.
\newblock An empirical study of the validity of the spearman-brown formula as
  applied to the purdue rating scale.
\newblock \emph{Journal of Educational Psychology}, 18(3):187.

\bibitem[{Sabou et~al.(2014)Sabou, Bontcheva, Derczynski, and
  Scharl}]{sabou2014}
Marta Sabou, Kalina Bontcheva, Leon Derczynski, and Arno Scharl. 2014.
\newblock Corpus annotation through crowdsourcing: Towards best practice
  guidelines.
\newblock In \emph{LREC}, pages 859--866. Citeseer.

\bibitem[{Sheshadri and Lease(2013)}]{sheshadri2013square}
Aashish Sheshadri and Matthew Lease. 2013.
\newblock Square: A benchmark for research on computing crowd consensus.
\newblock In \emph{Proceedings of the AAAI Conference on Human Computation and
  Crowdsourcing}, volume~1.

\bibitem[{Shrout and Fleiss(1979)}]{shrout1979}
Patrick~E Shrout and Joseph~L Fleiss. 1979.
\newblock Intraclass correlations: uses in assessing rater reliability.
\newblock \emph{Psychological bulletin}, 86(2):420.

\bibitem[{Snow et~al.(2008)Snow, O{'}Connor, Jurafsky, and Ng}]{snow2008}
Rion Snow, Brendan O{'}Connor, Daniel Jurafsky, and Andrew Ng. 2008.
\newblock \href {https://www.aclweb.org/anthology/D08-1027} {Cheap and fast
  {--} but is it good? evaluating non-expert annotations for natural language
  tasks}.
\newblock In \emph{Proceedings of the 2008 Conference on Empirical Methods in
  Natural Language Processing}, pages 254--263, Honolulu, Hawaii. Association
  for Computational Linguistics.

\bibitem[{Spearman(1910)}]{spearman1910}
Charles Spearman. 1910.
\newblock Correlation calculated from faulty data.
\newblock \emph{British Journal of Psychology, 1904-1920}, 3(3):271--295.

\bibitem[{Warrens(2017)}]{warrens2017}
Matthijs~J Warrens. 2017.
\newblock Transforming intraclass correlation coefficients with the
  spearman--brown formula.
\newblock \emph{Journal of clinical epidemiology}, 85:14--16.

\bibitem[{Wong et~al.(2021)Wong, Paritosh, and Aroyo}]{wong2021}
Ka~Wong, Praveen Paritosh, and Lora Aroyo. 2021.
\newblock \href {https://doi.org/10.18653/v1/2021.acl-long.548}
  {Cross-replication reliability - an empirical approach to interpreting
  inter-rater reliability}.
\newblock In \emph{Proceedings of the 59th Annual Meeting of the Association
  for Computational Linguistics and the 11th International Joint Conference on
  Natural Language Processing (Volume 1: Long Papers)}, pages 7053--7065,
  Online. Association for Computational Linguistics.

\bibitem[{Wulczyn et~al.(2017)Wulczyn, Thain, and Dixon}]{wulczyn2017ex}
Ellery Wulczyn, Nithum Thain, and Lucas Dixon. 2017.
\newblock Ex machina: Personal attacks seen at scale.
\newblock In \emph{Proceedings of the 26th international conference on world
  wide web}, pages 1391--1399.

\end{thebibliography}

\appendix
\section{Appendix on ICC($k$)}
\label{appendix}

\begin{figure}
  \centering
  \includegraphics[width=0.48\textwidth]{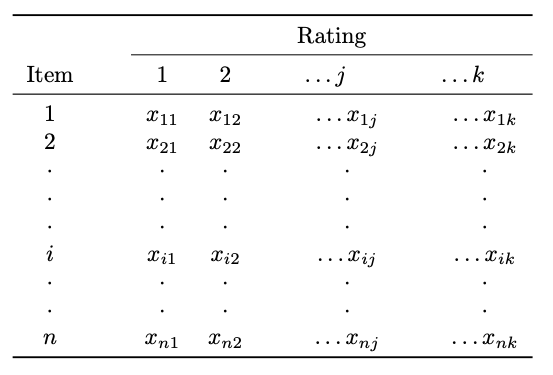}
  \caption{A convenient data matrix and notational system for data used in calculating intra-class correlation coefficients}
  \label{fig:data_matrix}
\end{figure}
ICC is a family of coefficients. It has slightly different formulations to accommodate different experimental designs. One of them, ICC($k$), quantifies the reliability of average ratings based on $k$ raters, where the raters are treated as interchangeable. We illustrate its close form calculation here. It is mainly re-expressing results from previous works on ICC calculation, such as \citet{liljequist2019} and \citet{mcgraw1996}.

ICC($k$) predicates on the one-way random effects model being the data generation process. The model takes the form
\[
x_{ij}=\mu+\phi_i+\epsilon_{ij},
\]
where $x_{ij}$ is the rating on item $i$ from rater $j$, $\mu$ is the grand mean, $\phi_i$ is the mean of item $i$, and $\epsilon_{ij}$ is a random perturbation term. Assume a data matrix with $n$ rows (item) and $k$ columns (raters) with no missing data, as one shown in Fig. \ref{fig:data_matrix}. Let
\[
\bar{x}_{\cdot\cdot}=\frac{1}{nk}\sum^k_{j=1}\sum^n_{i=1}x_{ij}
\]
be the sample grand mean, and 
\[
\bar{x}_{i\cdot}=\frac{1}{k}\sum^k_{j=1}x_{ij} 
\]
be the $i^\textrm{th}$ sample item mean. Let
\begin{align*}
SSW &= \sum^k_{j=1}\sum^n_{i=1}(x_{ij}-\bar{x}_{i\cdot})^2 \\
SSB &= k\sum^n_{i=1}(\bar{x}_{i\cdot}-\bar{x}_{\cdot\cdot})^2
\end{align*}
be respectively the sum of squares due to differences \textit{within} items and the sum of squares due to differences \textit{between} items. Then the estimator for the variance of $\epsilon$, $\sigma^2_\epsilon$, and the estimator for the variance of $\phi$, $\sigma^2_\phi$, are respectively
\begin{align*}
\hat{\sigma}^2_\epsilon &= \frac{SSW}{n(k-1)} \\
\hat{\sigma}^2_\phi &= \frac{SSB}{k(n-1)} - \frac{\hat{\sigma}^2_\epsilon}{k}.
\end{align*}
Then ICC($k$) can be computed as
\[
\frac{\hat{\sigma}^2_\phi}{\hat{\sigma}^2_\alpha+\hat{\sigma}^2_\epsilon/k}.
\]

If we apply the above formula to individual ratings, with $k=1$, the resulting reliability is known as inter-rater reliability. For any $k>1$, it is an instance of the \textit{k-rater reliability} proposed in this paper.

\end{document}